© Indian Academy of Sciences

# Machine Translation by Projecting Text into the Same Phonetic-Orthographic Space Using a Common Encoding

AMIT KUMAR[1,*], SHANTIPRIYA PARIDA[2], AJAY PRATAP[1], and ANIL KUMAR SINGH[1]

[1] Indian Institute of Technology (BHU), Varanasi, India
[2] Silo AI, Helsinki, Finland
e-mail: amitkumar.rs.cse17@iitbhu.ac.in (Corresponding author)

**Abstract.** The use of subword embedding has proved to be a major innovation in Neural Machine Translation (NMT). It helps NMT to learn better context vectors for Low Resource Languages (LRLs) so as to predict the target words by better modelling the morphologies of the two languages and also the morphosyntax transfer. Even so, their performance for translation in Indian language to Indian language scenario is still not as good as for resource-rich languages. One reason for this is the relative morphological richness of Indian languages, while another is that most of them fall into the extremely low resource or zero-shot categories. Since most major Indian languages use Indic or Brahmi origin scripts, the text written in them is highly phonetic in nature and phonetically similar in terms of abstract letters and their arrangements. We use these characteristics of Indian languages and their scripts to propose an approach based on common multilingual Latin-based encodings (WX notation) that take advantage of language similarity while addressing the morphological complexity issue in NMT. These multilingual Latin-based encodings in NMT, together with Byte Pair Embedding (BPE) allow us to better exploit their phonetic and orthographic as well as lexical similarities to improve the translation quality by projecting different but similar languages on the same orthographic-phonetic character space. We verify the proposed approach by demonstrating experiments on similar language pairs (Gujarati↔Hindi, Marathi↔Hindi, Nepali↔Hindi, Maithili↔Hindi, Punjabi↔Hindi, and Urdu↔Hindi) under low resource conditions. The proposed approach shows an improvement in a majority of cases, in one case as much as ∼10 BLEU points compared to baseline techniques for similar language pairs. We also get up to ∼1 BLEU points improvement on distant and zero-shot language pairs.

## 1 Introduction

Machine Translation (MT) has an interesting history in computation and research [20] with new paradigms being introduced over decades. MT achieved a watershed moment with the introduction of numerous algorithmic, architectural and training enhancements, such as Statistical Machine Translation (SMT) and Neural Machine Translation (NMT) [64]. SMT is a statistical-based MT paradigm, operating at the granularity of words and phrases, consisting of a translation model, a language model, and a decoder [17, 62, 66]. Further, the relatively recent success of deep neural networks has given us end-to-end variations of translation models such as recurrent NMT [21, 63], attention-based NMT, and self-attention-based Transformer [6].

There have been parallel and related developments in language models, such as Bidirectional Encoder Representations from Transformers (BERT) [22] and AL-BERT [23]. Another variant of this, mBART, has provided benchmark solutions in NMT as well [4]. However, training an effective and accurate MT system still requires a large amount of parallel corpus consisting of source and target language pairs. When we talk about low-resource languages, the first problem is to find a fair amount of parallel corpus, sometimes even monolingual corpus, which makes it challenging to create tools and applications for extremely poor resource languages. Creating a large parallel corpus for MT for each language pair that falls into the low resource category is an expensive, time-consuming, and labor-intensive task.

So, the solution to improve NMT in a low-resource context is to bootstrap the process by leveraging the morphological, structural, functional, and perhaps deep semantic features of such languages. Fortunately, for similar languages, it also is possible to exploit the similarities for better modeling of closely related languages. We need to focus on features that help the MT system better learn the close relationships between such languages. Conference on Machine Translation (WMT) has also conducted shared tasks for similar language translations from 2019 [24].

**Table 1**. Some details about the languages used in our experiments

| Languages | Family | Script | Word Order | Ergative | Place |
|-----------|--------|--------|------------|----------|-------|
| Hindi | | Devanagari | | Yes | Mainly North India |
| Gujarati | | Gujarati | | No | Mainly Gujarat |
| Marathi | | Balbodh version of Devanagari | | No | Mainly Maharashtra |
| Nepali | Indo-Aryan | Devanagari | SOV | Yes | Mainly Nepal |
| Maithili | | Devanagari | | No | Mainly Bihar and parts of Nepal |
| Punjabi | | Gurumukhi | | No | Mainly Punjab |
| Urdu | | Variant of Perso-Arabic | | No | Mainly North India |

When we talk about Indian languages, most languages except Hindi come under extremely low resource categories. Even Hindi is, from some points of view either a low or medium resource language [72, 73]. India being a country with rich linguistic diversity, there is a need for MT systems across the Indian (or South Asian) languages. India is also inhabited by a vast population who speak languages belonging to three prominent families, Indo-Aryan (a subfamily of Indo-European), Dravidian, and Tibeto-Burman, but due to very long contact and interactions, they have gone through a process of 'convergence', forming India as a linguistic area [25]. Due to this long term contact, there are more similarities among these languages than we would otherwise expect. In addition, significant fractions of their vocabularies, to varying degrees, have words originating in or borrowed from Sanskrit, Persian, Arabic, Turkish and English, among other languages.

For some of the major languages, and even for some of the 'regional' or 'minority languages' (since they were widely used for a long duration in the past for literary purposes), there are records available and there is a varying degree of well-developed tradition of at least (spoken) literary usage. However, only some languages, most of which are officially recognized, have some written tradition, particularly for non-literary prose. The rest have very little written data, or even if it is there, it is usually not in a machine-readable format. Therefore, they can be treated as extremely low or zero-resource languages. There is a need for development of MT systems for such languages, and the similarity between these languages helps in developing such MT systems.

In this article, we propose an approach based on leveraging the features of similar languages by simply, programmatically[1], converting them into an intermediate Latin-based multilingual notation. The notation that we use here is the commonly used WX-notation [26], which is often used in NLP tools and systems for Indian languages developed in India. This notation (like many other similar notations) can project all the Indic or Brahmi origin scripts [40], which have — in many cases — different Unicode blocks, into a common character space. Our intuition, is that this should help in capturing phonological, orthographic, and, to some extent, morphosyntactic similarities that will help a neural network-based model in better multilingual learning and translation across this languages [38, 39, 67]. We do this by using this WX-converted text to learn byte pair encoding-based embeddings. The effect of this is that the similar but different languages are projected onto the same orthographic-phonetic space [41], and hence also in the same common morphological and lexical space, allowing better modeling of multilingual relationships in the context of India as a linguistic area.

In addition, using WX has another benefit, even for a single script such as Devanagari. Brahmi-derived scripts have different symbols for dependent vowels (called *maatraas*) which modify a consonant and independent vowels (written as *aksharas*) which are pronounced as syllables. WX uses the same symbols for these two variants of the same vowel, while Unicode uses different codes and the scripts themselves use different graphical symbols.

After conversion to WX, we apply some of the state-of-the-art NMT techniques to build our MT systems. These NMT systems, such as the Transformer, should learn better the relationships between languages.

We select six pairs of similar languages: Gujarati (GU)↔Hindi (HI), Marathi (MR)↔Hindi (HI), Nepali (NE)↔Hindi (HI), Maithili (MAI)↔Hindi (HI), Punjabi (PA)↔Hindi (HI), and Urdu (UR)↔Hindi (HI). Table 1 contains some of the language features that help in figuring out how selected languages are similar to Hindi. For example, Hindi, Gujarati, Marathi, Nepali, Maithili, Punjabi, and Urdu belong to Indo-Aryan Language families, and all the selected languages except Punjabi and Urdu share a common Devanagari script. The word order of all the selected languages is mostly *Subject + Object + Verb*. Apart from this, all these languages share lexical similarities with Hindi in terms of common words derived from Sanskrit and other languages as mentioned earlier. Also, these languages have phonological similarities with Hindi. We also note that though Urdu and Hindi are linguistically almost the same language, yet due to the great divergence in their vocabularies in their written form, they have only a relatively small overlap in their corpus-based

---

vocabularies, albeit this overlap consists mainly of core words which form a major component of the linguistic identity of a language.

This papers is the first part of a series of three papers exploring and then extending the idea of using common phonetic-orthographic space for better NMT in the Indian context [68, 69]. The contributions of this paper are summarized as follows:

1. Propose a WX-based machine translation approach that leverages orthographic and phonological similarities between pairs of Indian languages.

2. Proposed approach achieves an improvement of *+0.01* to *+10* BLEU points compared to baseline state-of-the-art techniques for similar language pairs in most cases. We also get *+1* BLEU points improvement on distant and zero-shot language pairs.

The rest of the paper is organized as follows. Section 2 discusses closely related works. Section 3 describes some background and the NMT models that we extend or compare with. Section 4 describes the proposed approach in more detail. Section 5 discusses corpus statistics and experimental settings used to conduct the experiments. Results and ablation studies are reported in Sections 6 and 7, respectively. Finally, the paper is summarized in Section 8 and includes some directions for future work.

## 2 Related Works

This section briefly describes some of the related work (Table 2) on language similarity, morphological richness, statistical and neural models, and language pairs used as discussed below.

Although there had been work in the past, the recent sharper focus on machine translation for similar languages is also due to the shared tasks on this topic organized as part of the WMT conferences from 2019 to 2021. In [46], authors demonstrated that pre-training could help even when the language used for fine-tuning is absent during pre-training. In [47], authors experimented with attention-based recurrent neural network architecture (seq2seq) on HI↔MR and explored the use of different linguistic features like part-of-speech and morphological features, along with back translation for HI→MR and MR→HI machine translation. In [48], authors ensembled two Transformer models to try to allow the NMT system to learn the nuances of translation for low-resource language pairs by taking advantage of the fact that the source and target languages are written using the same script. In [49], authors' work relied on NMT with attention mechanism for the similar language translation in the WMT19 shared task in the context of NE↔HI language pair.

In [50], the authors conducted a series of experiments to address the challenges of translation between similar languages. Out of which, the authors developed one phrase-based SMT system and one NMT system using byte-pair embedding for the HI↔MR pair. In [51], authors used a Transformer-based NMT with *sentencepiece* for subword embedding on HI↔MR language pair [61]. In [52], authors used the Transformer-NMT for multilingual model training and evaluated the result on the HI↔MR pair. In [53], authors focused on incorporating monolingual data into NMT models with a back-translation approach. In [70], authors introduced NLP resources for 11 major Indian languages from two major language families. These resources include: large-scale sentence-level monolingual corpora, pre-trained word embeddings, pre-trained language models, and multiple NLU evaluation datasets. In [71], authors presented IndicBART, a multilingual, sequence-to-sequence pre-trained model focusing on 11 Indic languages and English. IndicBART utilized the orthographic similarity between Indic scripts to improve transfer learning between similar Indic languages.

### 2.1 *Shortcomings of existing works*

In most of the existing work on MT for related languages (e.g., [51], [52], [53]), authors have discussed improving the NMT models using extra monolingual corpora in addition to bi-lingual data. However, the proposed approach improves translation quality using only bilingual corpora with the help of WX-transliteration. The proposed approach reduces language complexity by transliterating the text to roman script and helps the NMT models to better learn the context information by exploiting language similarities. In this way, where applicable, it can complement the approaches which use extra monolingual data.

## 3 Background

This section provides some background on the recent most successful machine translation techniques. From vanilla NMT to more robust and advanced BART, a denoising autoencoder for pre-training sequence-to-sequence models, remarkable advances in NMT techniques have been made in a relatively short time.

### 3.1 *NMT*

Many of the NMT techniques use an encoder-decoder architecture based on neural networks that performs translation between language pairs. Numerous enhancements, toolkits, and open frameworks are available to train NMT models, such as OpenNMT. OpenNMT is one of the open-source NMT frameworks [2], used to model natural language tasks such as text summarization, tagging, and text generation. This toolkit is used for model architectures, feature representations, and source modalities in NMT research. Multilingual and zero-shot NMT have also been applied for NMT to achieve state-of-the-art results on different language pairs by

**Table 2**. Comparison of some existing work. ✓ and ✗ represent presence and absence of a particular feature, respectively.

| Paper | Similar Language | Reducing Morphological Complexity | Statistical | Neural | WX | Language Pair |
|---|---|---|---|---|---|---|
| [46] | ✓ | ✗ | ✗ | ✓ | ✗ | HI↔MR, ES↔PT |
| [47] | ✓ | ✗ | ✗ | ✓ | ✗ | HI↔MR |
| [48] | ✓ | ✗ | ✗ | ✓ | ✗ | HI↔MR |
| [49] | ✓ | ✗ | ✗ | ✓ | ✗ | NE↔HI |
| [50] | ✓ | ✗ | ✓ | ✓ | ✗ | HI↔MR |
| [51] | ✓ | ✗ | ✗ | ✓ | ✗ | HI↔MR |
| [52] | ✓ | ✗ | ✗ | ✓ | ✗ | HI↔MR |
| [53] | ✓ | ✗ | ✗ | ✓ | ✗ | ES↔PT, CS↔PL, NE↔HI |
| [70] | ✓ | ✗ | ✗ | ✓ | ✗ | 11 Indian languages |
| [71] | ✓ | ✗ | ✗ | ✓ | ✗ | 11 Indic languages and English |
| Proposed approach | ✓ | ✓ | ✗ | ✓ | ✓ | {GU,MR,NE,MAI,PA,UR}↔HI |

Note- HI: Hindi, MR: Marathi, ES: Spanish, PT: Portuguese, NE: Nepali, CS: Czech, PL:Polish, GU: Gujarati, MAI: Maithili, PA: Punjabi, UR: Urdu

using a single standard NMT model for multiple languages [5]. Furthermore, the introduction of 'attention' in NMT has drastically improved the results significantly [7], as for many other problems. As shown in Figure 1, NMT is an encoder-decoder sequence-based model consisting of recurrent neural network (RNN) units. The encoder consists of RNN units ($E_0$, $E_1$, $E_2$) and takes as input the embedding of words from sentences and produces the context vector ($\mathbf{C}$) as follows:

$$\mathbf{C} = Encoder(\mathbf{X_1}, \mathbf{X_2}, \mathbf{X_3}, ..., \mathbf{X_n}) \qquad (1)$$

where, $\{\mathbf{X_1}, \mathbf{X_2}, \mathbf{X_3}, ..., \mathbf{X_n}\}$ is the input source sequence.

The decoder consists of RNN units ($D_0$, $D_1$, $D_2$, $D_3$) and it decodes these context vectors into target sentences with an <END> (end of a sentence) symbol as follows:

$$Decoder(\mathbf{C}, \mathbf{Y_1}, \mathbf{Y_2}, \mathbf{Y_3}, ..., \mathbf{Y_n}) = \mathbf{Y'_1}, \mathbf{Y'_2}, \mathbf{Y'_3}, ..., \mathbf{Y'_m} \qquad (2)$$

where, $\{\mathbf{Y_1}, \mathbf{Y_2}, \mathbf{Y_3}, ..., \mathbf{Y_n}\}$ and $\{\mathbf{Y'_1}, \mathbf{Y'_2}, \mathbf{Y'_3}, ..., \mathbf{Y'_m}\}$ are target and predicted sequences, respectively.

### 3.2 *Transformer-based NMT*

The Transformer can be characterized by its breakthrough in combining five innovations elegantly in a single architecture. The first is the attention mechanism [6]. It maps a query and a set of key-value pairs to an output. A compatibility function of the query with the corresponding key computes the weights. The second extends the first by using multi-head self-attention. The third is the use of positional encoding in terms of relative positions, which allows it to learn temporal relationships and dependencies. The fourth is the use of masking, which has proved to be immensely effective in many other later models. The fifth is the use of residual connections. Together, the elegant combination of these innovations not only allows the model to learn much better models, but also obviates the need for recurrent units in the architecture, which in turn allows a great degree of parallelism during training the models. In other words, the Transformer model not only learns much better models, but does so in much less time during the training phase. Moreover, the problem of overfitting is also much less with the Transformer-based models.

There are numerous state-of-the-art results reported for machine translation systems using a Transformer. Currey and Heafield [8] incorporated syntax into the Transformer using a mixed encoder model and multi-task machine translation. Multi-head attention is one key feature of self-attention. Fixing the attention heads on the encoder side of the Transformer increases BLEU scores by up to 3 points in low-resource scenarios [9]. The most common attention functions are additive attention and dot product attention. Transformer generates the scaled dot-product attention as follows [6]:

$$\mathbf{attn_i} = softmax\left(\frac{\mathbf{Q_i}\mathbf{K_i}^T}{\sqrt{d_k}}\right)\mathbf{V_i} \qquad (3)$$

where, $\mathbf{Q_i}$, $\mathbf{K_i}$, $\mathbf{V_i}$ and $d_k$ are query, key, value and the dimension of the key, respectively.

### 3.3 *BART*

BART is a denoising autoencoder for pretraining sequence-to-sequence models [10]. It uses a standard Transformer-based NMT architecture to generalize BERT, GPT, and many other recent pre-training schemes. BART uses the standard Transformer architecture, except it modifies ReLU activation functions to GeLUs. Its mBART variation is a sequence-to-sequence denoising autoencoder pre-trained on monolingual corpora in multiple languages using the BART objective [4].

### 3.4 *Back-translation*

Back-translation is a method to prepare synthetic parallel corpus from a monolingual corpus for NMT [11]. In low-resource settings, back-translation can be a very effective method. Iterative back-translation is a further improvement [13]. It iterates over two back-translation systems multiple times.

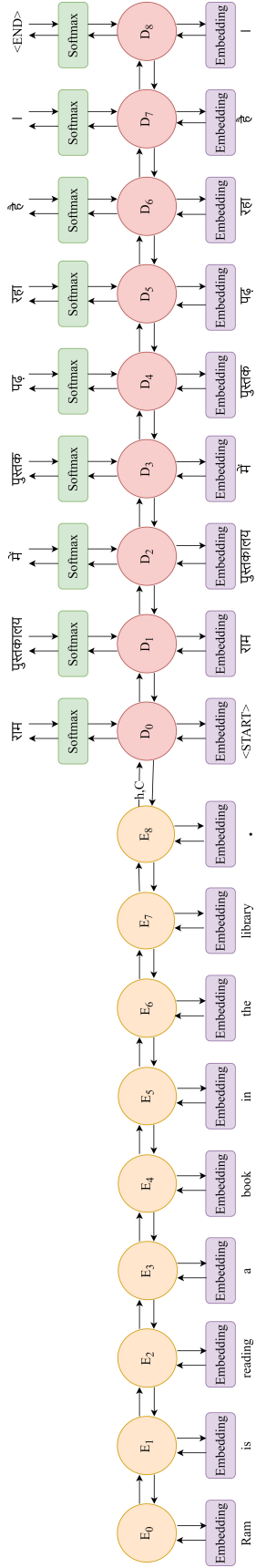

**Figure 1**. Vanilla NMT.

### 3.5 *Similar languages*

Similar languages refer to a group of languages that share common ancestry or extensive contact for an extended period, or both, with each other, leading them to exhibit structural and linguistic similarities even across language families. Examples of languages that share common ancestors are Indo-Aryan languages, Romance languages, and Slavic languages. Languages in contact for a long period lead to the convergence of linguistic features even if languages do not belong to common ancestors. Prolonged contact among languages could lead to the formation of linguistic areas or *sprachbunds*. Examples of such linguistic areas are the Indian subcontinent [25], the Balkan [42], and Standard Average European [43] linguistic areas.

Similarities between languages depend on various factors. Some of the factors are lexical similarity, structural correspondence, and morphological isomorphisms. Lexical similarity means that the languages share many words with similar forms (spelling/ pronunciation) and meaning, e.g. Sunday is written as रविवार (ravivAra) in Hindi and रिबिवार (rabiVra) in Bhojpuri (both are proximate and related Indo-Aryan languages). These lexically similar words could be cognates, lateral borrowings, or loan words from other languages. Structural correspondence means, for example, that languages have the same basic word order, viz. SOV (Subject-Object-Verb) or SVO (Subject-Verb-Object). Morphological isomorphisms refers to the one-to-one correspondence between inflectional affixes. While content words are borrowed or inherited across similar languages, function words are generally not lexically similar across languages. However, function words in related languages (whether suffixes or free words) tend to have a one-one correspondence to varying degrees and for various linguistic functions.

### 3.6 *Transformer-based NMT + Back-translation*

Guzmán et.al [3], in their work, first trained a Transformer on Nepali-English and Sinhala-English language pairs in both directions, and then they used the trained model to translate monolingual target language corpora to source languages. Finally, the source language sentence corpus was merged with generated source language sentences and was given as input to the Transformer for training and producing the translation.

### 4 Proposed Approach

To tackle the morphological richness related problems in NMT training for Indian languages and to be able work with very little resources, we propose a simple but effective approach for translating low-resource languages that are similar in features and behaviour.

The proposed approach consists of three modules: Text Encoder, Model Trainer, and Text Decoder (Figure 2), as discussed in the following section.

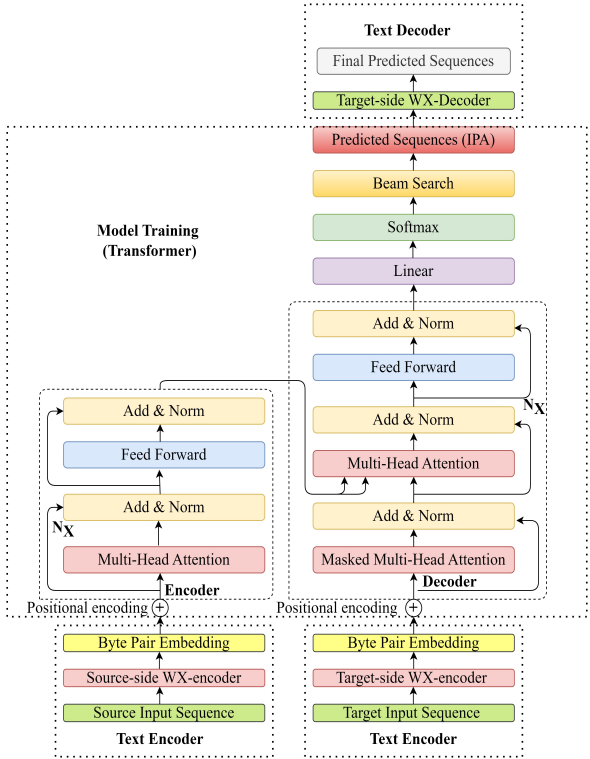

**Figure 2**. Proposed architecture.

### 4.1 *Text Encoder*

The proposed model first encodes the source and target corpora of parallel languages into an intermediate representation, the WX-notation[2] [1]. The primary reason behind encoding the source and target language corpora into WX-notation is to encode different languages with the same or different scripts into a common representation by projecting them onto a common phonetic-orthographic character space so that BPE can be linguistically better informed. WX-notation is a transliteration scheme for representing Indian languages in ASCII format, and as described earlier, it has many advantaged as an intermediate representation, even compared to using Devanagari or any other single Brahmi-based script. It implicitly helps the Transformer encoder model more cognates, loan words, and morphologically similar words between the languages, as well as model other kinds of similarities for better translation.

### 4.2 *Model Training*

The intermediate representation of the source language text is passed to the Transformer encoder. The Transformer encoder-decoder model learns the relationship between languages. We have used the SentencePiece[3] li-brary for tokenization of the text. SentencePiece is used as a pre-processing task for the WX-encoded source-target text in the concerned language pair. Sentence-Piece is a language-independent sub-word tokenizer and detokenizer designed for Neural-based text processing, including neural machine translation. It implements two subword segmentation algorithms, Byte-Pair Encoding (BPE) and unigram language model, with direct training from raw sentences [33, 34]. Therefore, it already indirectly, to some extent, provides cognates, loan words, and morphologically similar words to the Transformer, and our prior conversion to WX allows it to do so better. It may be noted that the approach is generalizable to other multilingual transliteration notations, perhaps even to IPA[4,5], which is almost truly phonetic notation for written text.

### 4.3 *Text Decoder*

After convergence of the training algorithm, the WX-encoded generated target sentences are decoded back to the plain text format to evaluate the model.

## 5 Corpus and Experimental Settings

In this section, we discuss the corpus statistics and experimental settings we used for our experiment.

### 5.1 *Corpus description*

We evaluate the proposed model in an extremely low-resource scenario on the mutually similar languages which we selected for our experiments. These are Hindi (HI), Gujarati (GU), Marathi (MR), Nepali (NE), Maithili (MAI), Punjabi (PA), Urdu (UR), Bhojpuri (BHO), Magahi (MAG), Malayalam (ML), Tamil (TA) and Telgu (TE). We perform experiments on the following language pairs involving Hindi: GU↔HI, NE↔HI, MR↔HI, MAI↔HI, PA↔HI, and UR↔HI. Parallel corpora of GU↔HI, ML↔HI, TA↔HI, and TE↔HI for training, testing, and validation are downloaded from CVIT-PIB [14]. MR↔HI parallel corpus is collected from WMT 2020 shared tasks[6]. NE↔HI language pair corpus is made up of those collected from WMT 2019 shared tasks [7], Opus [8], and TDIL [9] repositories. We use a monolingual corpus of Gujarati, Hindi, and Marathi for similarity computation in section 5.1 from the PM India dataset described in [15]. The rest of the monolingual corpora are collected from the Opus collection for similarity computation in section 5.1 [29]. We use Sentence-Piece [45] to pre-process the source and target sentences.

---

**Table 3**. Corpus Statistics showing the number of training, validation, and test sentences for each domain

| Lang-Pairs | Train | Validation | Test | Domain |
|---|---|---|---|---|
| GU↔HI | 15784 | 1000 | 1973 | PM India |
| NE↔HI | 136991 | 3000 | 3000 | WMT 2019 corpus, Agriculture, Entertainment, Bible |
| MR↔HI | 43274 | 1000 | 1411 | News, PM India, Indic WordNet |
| PA↔HI | 225576 | 7199 | 7200 | GNOME, KDE4, Ubuntu, wikimedia, TED2020 |
| MAI↔HI | 93136 | 2972 | 2973 | GNOME, KDE4, wikimedia, Ubuntu |
| UR↔HI | 108176 | 3452 | 3453 | Tanzil, GNOME, KDE4, wikimedia, Ubuntu |
| ML↔HI | 17333 | 500 | 500 | PM India |
| TA↔HI | 43538 | 500 | 500 | PM India |
| TE↔HI | 2584 | 500 | 500 | PM India |
| BHO↔HI | 0 | 500 | 500 | Movie subtitles, Literature, News |
| MAG↔HI | 0 | 500 | 500 | Movie subtitles, Literature, News |

Note: HI: Hindi, MR: Marathi, NE: Nepali, GU: Gujarati, MAI: Maithili, PA: Punjabi, UR: Urdu, ML: Malayalam, TA: Tamil, TE: Telgu, BHO: Bhojpuri, MAG: Magahi

We use 5K merge operations to learn BPE with the SentencePiece model and restrict the source and target vocabularies to at most 5K tokens. There are some places where code-switching occurs in the employed dataset. The WX-transliteration tool ignores code-switched data and keeps it in the datasets as it is.

## 5.2 *Training details*

### 5.2.1 *Proposed approach*

We use the WX-notation tool [10] for transliterating the text and the fairseq [11] [18] toolkit, which is a sequence modelling toolkit, to train the Transformer. We use five encoder and decoder layers. The encoder and decoder embedding dimensions are set to 512. Feed-forward encoding and decoding embedding dimensions are set to 2048. The number of an encoder and decoder attention heads is set to 2. The dropout, the attention dropout, and the ReLU dropout are set to 0.4, 0.2, and 0.2, respectively. The weight decay is set at 0.0001, and the label smoothing is set to 0.2. We use the Adam optimizer, with $\beta_1$ and $\beta_2$ set to 0.9 and 0.98. The learning rate schedule is inverse square root, with an initial learning rate of 1e-3 and a minimum learning rate of 1e-9. The maximum number of tokens used is set to 4000. The maximum number of epochs for training is set to 100. We use a beam size equal to 5 for generating data using the test set.

### 5.2.2 *Guzmán et al. [3]*

In Guzmán et al. [3], authors have demonstrated the experiments on extremely low resource languages using Transformer. Our proposed approach is based on the Transformer described in Guzmán et al. [3] with the addition of two extra modules, Text Encoder and Text Decoder. We use the Transformer model described in

Guzmán et al. [3] as a reasonably high baseline to compare the proposed approach without the intermediate representation of the WX-notation for Indian languages. The projection to WX could be used for any other NMT approach as well that uses a subword embedding.

### 5.2.3 *SMT*

We use Moses [12], an open-source toolkit to train SMT [54]. For obtaining the phrase/word alignments from parallel corpora, we use *GIZA++* [55]. A 5-gram KenLM language model is used for training [56]. The parameters are tuned on the validation set using *MERT* and tested with a test set [57].

## 6 Results and Analysis

We compare the proposed approach with the Moses-based SMT and the Transformer-based NMT model [3], where the latter is used as the baseline for NMT. We use six evaluation metrics, BLEU [13] [12], LEBLEU [58], WupLeBleu [59], TER [31], WER, and chrF2 [30] for better comparison of the proposed approach. We see from Tables 4 and 5 that the proposed approach improves upon the baseline for most of the pairs.

BLEU score, although a simple metric based on comparison of *n*-grams, is a standard metric accepted by *NLP* researchers to obtain the accuracy of predicted translated outputs compared to the human-translated reference sentences. This is because it has been observed that the value of the BLEU score correlates well with human-judged quality of translations. The formula for the BLEU score is as follows [12]:

$$BLEU = min\left(1, \frac{output\_length}{reference\_length}\right)\left(\prod_{i=1}^{4} precision_i\right), \quad (4)$$

**Table 4.** Experiment results (BLEU, chrF2, and TER scores).

| Languages(xx) | BLEU | | chrF2 | | TER | |
|---|---|---|---|---|---|---|
| | **XX→HI** | | | | | |
| | Guzmán et.al [3] | Proposed | Guzmán et.al [3] | Proposed | Guzmán et.al [3] | Proposed |
| GU | 33.14 | **33.15** | **58** | 57 | **0.541** | 0.548 |
| NE | 30.51 | **41.97** | 46 | **49** | 0.658 | **0.652** |
| MR | 16.87 | **22.37** | 43 | **44** | **0.707** | 0.709 |
| PA | 78.56 | **81.05** | 82 | **82** | 0.220 | **0.216** |
| UR | 28.74 | **30.08** | 45 | **45** | 0.668 | **0.657** |
| MAI | 79.49 | **81.80** | **82** | 81 | **0.242** | 0.251 |
| | **HI→XX** | | | | | |
| | Guzmán et.al [3] | Proposed | Guzmán et.al [3] | Proposed | Guzmán et.al [3] | Proposed |
| GU | 25.47 | **25.82** | 56 | **56** | **0.616** | 0.619 |
| NE | 32.89 | **43.52** | 50 | **51** | **0.630** | 0.637 |
| MR | 14.05 | **14.76** | 41 | **44** | 0.789 | **0.762** |
| PA | 80.01 | **81.87** | 83 | **84** | 0.206 | **0.203** |
| UR | 22.74 | **24.35** | 46 | **47** | 0.597 | **0.596** |
| MAI | **86.58** | 83.82 | **89** | 86 | **0.148** | 0.168 |

**Table 5.** LEBLEU, WupLeBleu and WER scores.

| Languages(xx) | LEBLEU | | WupLeBLEU | | WER | |
|---|---|---|---|---|---|---|
| | **XX→HI** | | | | | |
| | Guzmán et.al [3] | Proposed | Guzmán et.al [3] | Proposed | Guzmán et.al [3] | Proposed |
| GU | **0.663** | 0.657 | **0.663** | 0.657 | 66.77 | **66.29** |
| NE | 0.543 | **0.547** | 0.543 | **0.547** | **66.99** | 67.71 |
| MR | 0.495 | **0.541** | 0.495 | **0.541** | **72.78** | 73.36 |
| PA | 0.853 | **0.853** | 0.853 | **0.853** | 22.29 | **21.83** |
| UR | 0.564 | **0.566** | 0.564 | **0.566** | 68.34 | **67.20** |
| MAI | **0.865** | 0.851 | **0.865** | 0.851 | **24.34** | 25.23 |
| | **HI→XX** | | | | | |
| | Guzmán et.al [3] | Proposed | Guzmán et.al [3] | Proposed | Guzmán et.al [3] | Proposed |
| GU | 0.622 | **0.623** | 0.622 | **0.623** | **73.11** | 73.33 |
| NE | **0.547** | 0.519 | **0.547** | 0.519 | **63.41** | 65.31 |
| MR | **0.485** | 0.454 | **0.485** | 0.454 | 80.10 | **77.46** |
| PA | 0.858 | **0.865** | 0.858 | **0.865** | 20.88 | **20.57** |
| UR | 0.619 | **0.629** | 0.619 | **0.629** | 62.35 | **62.27** |
| MAI | **0.916** | 0.908 | **0.916** | 0.908 | **14.83** | 16.89 |

where the *output_length* and the *reference_length* are the lengths of the predicted sentences and the reference sentences, respectively.

We also perform a comparison between SMT without WX-transliteration and SMT with it. These two sets of results are also compared with the proposed approach as shown in Table 6. In the case of SMT also we can easily note that the performance improves in most cases by using WX as the intermediate notation, even though SMT is not using subword embeddings.

We also present some basic analysis of the scores as shown in Tables 4 and 5. We use corpus-based language relatedness and complexity measures for further analysis for this purpose in the next section.

### 6.1 *Similarity between languages*

Since there are no definitive methods to judge the similarity between two languages, we use the following techniques to compute the similarity between the languages:

#### 6.1.1 *SSNGLMScore*

We use character-level *n*-gram language models based SSNGLMScore to measure the relatedness between languages [28, 32]. SSNGLMScore is computed as follows:

$$S_{sl,tl} = \sum_{tl=1}^{m} p_{sl,tl}(w_n|w_1^{n-1}), \qquad (5)$$

where $S$ stands for Scaled Sum of *n*-gram language model scores.

$$MS_{sl,tl} = \frac{S_{sl,tl} - \min(S_{SL,TL})}{\max(S_{SL,TL}) - \min(S_{SL,TL})}, \qquad (6)$$

where, *sl* and *tl* represent the source language and the target language, respectively. Moreover, *sl* ∈ SL(Gujarati, Marathi, Maithili, Nepali, Urdu, Punjabi, Hindi, Malayalam, Tamil, Telugu, Bhojpuri, Magahi) and *m* is the total number of sentences in the target language *tl* ∈ TL(Gujarati, Marathi, Maithili, Nepali, Urdu, Punjabi, Hindi, Malayalam, Tamil, Telugu, Bhojpuri, Magahi). We train the language model using a 6-gram character-level KenLM model on the source monolingual corpus

**Table 6**. BLEU score-based comparison of SMT, SMT + WX and the proposed approaches.

| Languages(xx) | BLEU | | |
|---|---|---|---|
| | XX→HI | | |
| | SMT | SMT + WX | Proposed |
| GU | **43.49** | 30.69 | 33.15 |
| NE | 40.14 | **53.21** | 41.97 |
| MR | **7.41** | 1.46 | 22.37 |
| PA | 68.34 | **71.22** | 81.05 |
| UR | 19.21 | **21.84** | 30.08 |
| MAI | 79.56 | **81.46** | 81.80 |
| | HI→XX | | |
| | SMT | SMT + WX | Proposed |
| GU | **39.20** | 25.89 | 25.82 |
| NE | 40.21 | **54.84** | 43.52 |
| MR | **7.36** | 1.48 | 14.76 |
| PA | 67.21 | **70.64** | 81.87 |
| UR | 18.24 | **18.41** | 24.35 |
| MAI | 79.12 | **83.06** | 83.82 |

(*sl*). Each language model is tested on target language (*tl*), and the scores are reported.

Table 7 lists the cross-lingual similarity scores of Hindi, Gujarati, Marathi, Nepali, Maithili, Punjabi, Malayalam, Tamil, Telugu, Bhojpuri, Magahi, and Urdu with each other. Based on SSNGLMScore, Bhojpuri, Maithili and Magahi are the closest to Hindi, which matches linguistic knowledge about them, whereas Urdu seems to as far from Hindi as Malayalam and more than Telugu. The reasons Urdu is far from Hindi is partly that Urdu is written in a different kind of script from Hindi which does not have a straightforward mapping to WX, but mainly because, though grammatically almost identical, the two use very different vocabularies in written and formal forms. Maithili is also the second official language of Nepal and is also highly similar to Nepali, perhaps due to prolonged close contact. What is more surprising is that the similarity between Urdu and Nepali is relatively high, whereas that between Urdu and Hindi is among the lowest. This could be because of the nature of the corpus. Going through Tables 4 and 5, we find that there is an improvement in every metric except WER and TER in a majority of cases when we apply the proposed method on the translation direction from Maithili, Gujarati, Marathi, Nepali, Punjabi, and Urdu to Hindi. This observation allows us to assert that the proposed approach improves performance for translation between similar languages. Thus, even though the similarity measure we used mixes different kinds of similarities, it is suitable for our purposes because our method is based on sub-word and multilingual modelling.

We also see a gain of +1.34 BLEU points on Hindi to Urdu despite Urdu being far away from the rest of the language pairs in terms of the similarity score we used. There is a considerable improvement of +11.46 BLEU points on HI→NE and +10.63 BLEU points on NE→HI language pairs.

### 6.1.2 *char-BLEU, TER and chrF2*

To better understand the slight fall in BLEU points despite the similarity for MAI → HI and large increment in the case of NE↔Hi (where Nepali and Maithili are known to be close), we also compute similarity by applying char-BLEU [44], chrF2, and TER on a training dataset of all language pairs. The reason behind using char-BLEU and chrF2 for similarity is that since they are character-based metrics, there is a greater chance of covering the morphological aspects. Before calculating the char-BLEU, the TER, and the chrF2 evaluation metrics, data must be in the same script to evaluate the score. So, we convert the corpus from UTF-8 to WX-notation. Table 8 contains the char-BLEU score of language pairs, whereas Table 9 contains the TER and chrF2 scores of each language pair. We see Table 8 and 9 and find out that HI and MAI are still more similar compared to other pairs. We can only hypothesize the reason being that this is due to the nature of the data that we have used.

### 6.2 *Analysis on language complexity*

#### 6.2.1 *Morphological complexity*

Since Indian languages are morphologically rich, machine translation systems based on word tokens have difficulty with them. Therefore, we also tried to relate the results obtained with estimates of such complexity obtained from character-level entropy. It is reasonable to assume that the greater the character-level entropy, the more morphologically complex a language is likely to be.

Character-level entropy   We used Character-level word entropy to estimate morphological redundancy, following Bharati et al.[74] and Bentz and Alikaniotis 2016 [35].

A "word" is defined in our experiments as a space-separated token, i.e., a string of alphanumeric Unicode characters delimited by white spaces. The average information content of character types for words is then calculated in terms of Shannon entropy [36]:

$$H(T) = -\sum_{i=1}^{V} p(c_i) \log_2(p(c_i)) \qquad (7)$$

where $V$ is number of characters ($c_i$) in a word.

Table 10 lists the word (unigram) entropy of languages at character level, which indirectly represents languages' lexical richness, i.e., how complex – in terms of characters they are made up of – word forms are. Since we compute the unigram entropy based on characters, we can say that lexical richness also indicates morphological complexity, both derivational and inflectional. Based on the corpus-based word entropy values, it appears that Hindi is more morphologically complex than the other six languages. However, this may be more of derivational complexity rather than inflectional

**Table 7.** Similarity between languages using SSNGLMScore

| Model | BHO | GU | HI | MAG | MAI | ML | MR | NE | PA | TA | TE | UR |
|---|---|---|---|---|---|---|---|---|---|---|---|---|
| **BHO** | - | 0.5659 | 0.6725 | 0.6997 | 0.7235 | 0.4090 | 0.5687 | 0.4979 | 0.4580 | 0.3233 | 0.5057 | 0.4237 |
| **GU** | - | - | 0.5483 | 0.5642 | 0.6449 | 0.3727 | 0.5411 | 0.3868 | 0.3408 | 0.2531 | 0.4578 | 0.3787 |
| **HI** | - | - | - | 0.6331 | 0.6598 | 0.3536 | 0.5717 | 0.4181 | 0.4046 | 0.2564 | 0.4567 | 0.3670 |
| **MAG** | - | - | - | - | 0.7762 | 0.4414 | 0.5724 | 0.5671 | 0.4827 | 0.3736 | 0.5248 | 0.5245 |
| **MAI** | - | - | - | - | - | 0.5833 | 0.6496 | 0.6968 | 0.5734 | 0.5453 | 0.6435 | 0.7040 |
| **ML** | - | - | - | - | - | - | 0.3736 | 0.3388 | 0.1968 | 0.3792 | 0.4507 | 0.2759 |
| **MR** | - | - | - | - | - | - | - | 0.4023 | 0.3496 | 0.2637 | 0.4771 | 0.3498 |
| **NE** | - | - | - | - | - | - | - | - | 0.2661 | 0.2784 | 0.3985 | 0.4354 |
| **PA** | - | - | - | - | - | - | - | - | - | 0.1449 | 0.2718 | 0.2938 |
| **TA** | - | - | - | - | - | - | - | - | - | - | 0.2972 | 0.2641 |
| **TE** | - | - | - | - | - | - | - | - | - | - | - | 0.3493 |
| **UR** | - | - | - | - | - | - | - | - | - | - | - | - |

**Table 8.** char-BLEU score on the training data

| Languages | char-BLEU |
|---|---|
| Gujarati↔Hindi | 47.29 |
| Marathi↔Hindi | 35.05 |
| Nepali↔Hindi | 40.53 |
| Maithili↔Hindi | 66.70 |
| Punjabi↔Hindi | 37.17 |
| Urdu↔Hindi | 8.61 |

Note: Applying char-BLEU score on the training data of both the languages of the pair

complexity, as Hindi is relatively simpler in terms of inflectional morphology. The high derivational complexity of Hindi is because it is the official language of India and is more standardized than most other Indian languages. It, therefore, has borrowed and coined a large number of complicated words and technical terms, whether from Persian or Sanskrit or English. This adds a great deal to the derivational complexity of written formal Hindi, compared to commonly spoken Hindi. At least, this is our hypothesis based on the similarity and complexity results.

We also find that our approach shows a considerable improvement of about more than 10 BLEU points in both directions for the Hindi-Nepali language pair, i.e., NE→HI and HI→NE. Such improvement may be attributed to the effect caused by projecting to a common multilingual orthographic-phonetic notation, that is, WX. This probably helps the Transformer learn the context between languages better with the help of a sentence piece tokenizer.

In Tables 11, 12 and 13, we present the values of word entropy and redundancy at character level. These tables show that the entropy increases when converting to WX and redundancy decreases. This is evidence of the fact that the project to a common orthographic and phonetic space causes the entropy to increase and redundancy to decrease, thus allowing more compact representations to be learnt from the data after conversion to WX in our case.

### 6.2.2 Syntactic complexity

**Perplexity** Perplexity ($PP$) of a language can be seen as a weighted average of the reciprocal of its branching factor [28]. Branching factor is the number of possible words that can succeed any given word based on the context. Therefore, perplexity – as a kind of the mean branching factor – is a mean representative of the possible succeeding words given a word. Thus, it can be seen as a rough measure of the syntactic complexity. If the model is a good enough representation of the true distribution for the language, then the $PP$ value will actually indicate syntactic complexity.

To estimate distances of other languages from Hindi using perplexity, we trained the perplexity model on the Hindi corpus and tested it on the corpora of other languages.

$$PP(C) = \sqrt[w]{\frac{1}{P(S_1, S_2, S_3, ..., S_n)}} \qquad (8)$$

where corpus $C$ contains $n$ sentences with $W$ words.

Table 14 and 15 contain the assymmetric and symetric perplexity — average of the two translation directions — values between the concerned language pairs and indicate their distances from Hindi based on character-level language model. Pairs having higher perplexity scores means the concerned languages are more distant. We see language pairs Urdu and Hindi have more perplexity scores. This is mostly because these two languages, though almost identical in spoken form and in terms of core syntax and core vocabulary, use very different extended vocabularies for written and formal purposes, besides using very different writing systems. Standard written Urdu uses Persian, Arabic, and Turkish words heavily, whether adapted phonologically or not.

Given the small amounts of data, it is not surprising that the values of perplexity are different in the two translation directions.

Similarly, standard and written Hindi uses words much more heavily derived or borrowed or even coined

**Table 9.** TER and chrF2 scores on the training data

| Languages | $GU \rightarrow HI$ | $MR \rightarrow HI$ | $NE \rightarrow HI$ | $MAI \rightarrow HI$ | $PA \rightarrow HI$ | $UR \rightarrow HI$ |
|---|---|---|---|---|---|---|
| TER | 1.066 | 1.300 | 1.052 | 0.610 | 0.988 | 1.093 |
| chrF2 | 38 | 29 | 34 | 65 | 32 | 12 |
| Languages | $HI \rightarrow GU$ | $HI \rightarrow MR$ | $HI \rightarrow NE$ | $HI \rightarrow MAI$ | $HI \rightarrow PA$ | $HI \rightarrow UR$ |
| TER | 0.884 | 0.940 | 0.887 | 0.555 | 0.906 | 1.044 |
| chrF2 | 39 | 29 | 36 | 62 | 30 | 10 |

Note: Applying TER and chrF2 scores on the training data of both the languages of a pair

**Table 10.** Character-based entropy of languages with or without applying WX-notation

| Languages | Character Entropy | Character Entropy* | Difference |
|---|---|---|---|
| Gujarati | 5.0368 | 3.7454 | 1.2914 |
| Marathi | 5.0220 | 3.6846 | 1.3374 |
| Nepali | 4.6722 | 3.5770 | 1.0952 |
| Maithili | 5.1159 | 3.9162 | 1.1997 |
| Punjabi | 5.0834 | 3.7932 | 1.2902 |
| Urdu | 4.8821 | 4.1198 | 0.7623 |
| Hindi | 5.2195 | 3.7974 | 1.4221 |

* After applying WX-notation

from Sanskrit. Despite higher perplexity between these two languages, our approach gives a *+2* increment in the BLEU score, probably because the common core syntax and core vocabulary manifest themselves in every phrase or sentence and thus have higher probabilistic weight. They are, in fact, completely mutually intelligible in the spoken forms and partly in the written form. There are also a lot of Indians who can comfortably read and understand both these languages, even in their standard, written, and literary forms. The use of WX perhaps allows the models to exploit the core similarities better.

## 7  Ablation Study

This section discusses ablation studies conducted using the proposed method on distant and zero-shot language pairs and back-translation.

### 7.1  *Analysis of the proposed approach on more distant language pairs*

To see whether and to what extent our approach generalizes to more distant language pairs, we also analyze the performance of the proposed approach on (ML↔HI, TA↔HI, and TE↔HI). Malayalam, Tamil, and Telugu belong the Dravidian family, and Hindi is from the Indo-Aryan family. We note that translating between these three Dravidian languages and Hindi still leads to improvement, considering both chrF2 and BLEU scores. The results are shown in Table 16.

### 7.2  *Unsupervised settings*

We also demonstrate the proposed approach under unsupervised scenarios on zero-shot language pairs, Bhojpuri-Hindi and Magahi-Hindi, for which no parallel train-ing corpora is available. The validation datasets for zero-shot experiments are collected from LoResMT 2020 shared tasks[14]. For training the model, we use NE↔HI language pairs and use language transfer on zero-shot pairs to evaluate the model on validation datasets. The reason behind using NE↔HI language pairs for training the model in unsupervised experiments on Bhojpuri-Hindi and Magahi-Hindi is the higher similarity between NE↔HI language pairs with both Bhojpuri-Hindi and Magahi-Hindi zero-shot language pairs based on [65]. The results are shown in Table 17, demonstrating the improvement in unsupervised settings also.

### 7.3  *Back-translation*

Finally we report results on using the approach along with Back-Translation, which has been shown to benefit machine translation for very low resource languages. We selected Gujarati and Hindi language pairs for performing Back-Translation (BT) with the proposed approach. With Back-Translation also, the proposed approach shows an improvement of BLEU point *+0.97* on HI→GU and *+1.36* on GU→HI language pairs, as shown in Table 18.

## 8  Conclusion and Future Scope

In this work, we have proposed a simple but effective MT system approach by encoding the source and target script into an intermediate representation, WX-notation, that helps the models to be learnt in a common phonetic and orthographic space. This language projection reduces the surface complexity of the algorithm and allows the neural network to better model the relationships between languages to provide an improved translation. Further, we have investigated these results by estimating the similarities and complexities of language pairs and individual languages to verify that our results are consistent and agree with the intuitively known facts about the closeness or distances between various language pairs. Moreover, this approach works well under unsupervised settings and works fine for some distant language pairs. The proposed approach improves baseline approaches by *0.01* BLEU points to *11.46* BLEU points. The proposed approach has some limitations and boundary conditions. First, it requires a common transliteration script, which may not be available for all morphologically rich languages. Second, it is only applicable to Indian languages. Third, we can see from Table 16 that performance on distant language pairs falls short of expectations.

---

[14]https://sites.google.com/view/loresmt

**Table 11.** Entropy computed on Vocabulary

| Language | Complete corpus | | | | | | Restricted corpus | | | | | |
|---|---|---|---|---|---|---|---|---|---|---|---|---|
| | Without WX | | | With WX | | | Without WX | | | With WX | | |
| | Max | Median | Average | Max | Median | Average | Max | Median | Average | Max | Median | Average |
| **HI** | 3.1674 | 0.5897 | 0.6196 | 4.9433 | 1.2484 | 1.3148 | 3.1623 | 0.5929 | 0.6230 | 4.9414 | 1.2495 | 1.3158 |
| **GU** | 6.4712 | 0.8113 | 0.8389 | 17.9337 | 1.4677 | 1.5157 | 6.4735 | 0.8128 | 0.8410 | 22.2253 | 1.4681 | 1.5163 |
| **NE** | 3.0311 | 0.8008 | 0.8287 | 6.6845 | 1.4327 | 1.4835 | 1.8080 | 0.5350 | 0.5636 | 4.7487 | 1.1262 | 1.1575 |
| **MR** | 3.7534 | 0.5982 | 0.6281 | 7.7372 | 1.2331 | 1.2995 | 3.5845 | 0.8049 | 0.8459 | 7.7400 | 1.2130 | 1.2734 |
| **PA** | 2.2077 | 0.5778 | 0.6048 | 8.9978 | 1.0349 | 1.1105 | 2.1662 | 0.5500 | 0.5753 | 13.5759 | 0.9644 | 1.0405 |
| **UR** | 2.8580 | 0.6484 | 0.6786 | 3.092 | 0.7748 | 0.8088 | 2.2477 | 0.6282 | 0.6574 | 3.3297 | 0.7523 | 0.7828 |
| **MAI** | 2.0163 | 0.5097 | 0.5326 | 4.3135 | 1.0904 | 1.1432 | 1.6417 | 0.4773 | 0.5003 | 3.8923 | 1.0401 | 1.0888 |

**Table 12.** Redundancy

| Languages | Complete corpus | | Restricted corpus | |
|---|---|---|---|---|
| | Without WX | WX | Without WX | WX |
| HI | 0.8955 | 0.7693 | 0.8949 | 0.7691 |
| GU | 0.8606 | 0.7401 | 0.8603 | 0.7400 |
| NE | 0.8806 | 0.7866 | 0.9111 | 0.8147 |
| MR | 0.9050 | 0.7993 | 0.8610 | 0.7807 |
| PA | 0.9186 | 0.8502 | 0.9194 | 0.8554 |
| UR | 0.8941 | 0.8741 | 0.8968 | 0.8750 |
| MAI | 0.9125 | 0.8121 | 0.9172 | 0.8171 |

In the future, we plan to extend this approach to the various ways described below:

a. **Multilingual NMT system:** Since the proposed approach transforms all the Indian language scripts into a common notation called WX, this conversion favours the subword embeddings to work as character embedding. It may be, therefore, more beneficial to implement this approach in the multilingual system(s) for all Indian languages.

b. **BART, MBART, and other representations:** We tried the MBART-based translation of Gujarati to Hindi and Hindi to Gujarati, and the results are worse than a vanilla transformer. So, we plan to extend the proposed approach to more representations like BART, MBART, and other state-of-the-art representation techniques for Deep Learning.

c. **Dravidian languages and the rest of the Indo-Aryan language family:** We also plan to extend the proposed approach to the Dravidian language family and the rest of the Indo-Aryan languages.

**Table 13.** Entropy and Redundancy

| Language pair | Without WX | | | | With WX | | | |
|---|---|---|---|---|---|---|---|---|
| | Maximum Entropy | Median Entropy | Average Entropy | Redundancy | Maximum Entropy | Median Entropy | Average Entropy | Redundancy |
| GU-HI | 4.8292 | 0.43224 | 0.4985 | 0.9279 | 17.7731 | 1.3958 | 1.4509 | 0.7512 |
| NE-HI | 3.0273 | 0.7414 | 0.7725 | 0.8948 | 7.1454 | 1.3561 | 1.4126 | 0.7988 |
| MR-HI | 3.7557 | 0.6003 | 0.6303 | 0.9047 | 7.7342 | 1.2309 | 1.2977 | 0.7995 |
| PA-HI | 1.6642 | 0.3359 | 0.3510 | 0.9543 | 9.0232 | 1.1199 | 1.1843 | 0.8414 |
| UR-HI | 1.9841 | 0.3547 | 0.3864 | 0.9489 | 4.0133 | 0.7928 | 0.8472 | 0.8783 |
| MAI-HI | 2.0483 | 0.5340 | 0.5555 | 0.9096 | 6.8270 | 1.1097 | 1.1656 | 0.8091 |

**Table 14.** Cross-lingual distance between languages after applying character-level language model using perplexity-based score (Unnormalized on language directions)

| Language | BHO | GU | HI | MAG | MAI | ML | MR | NE | PA | TA | TE | UR |
|---|---|---|---|---|---|---|---|---|---|---|---|---|
| BHO | 0.0010 | 0.0443 | 0.0280 | 0.0290 | 0.0617 | 0.1006 | 0.0418 | 0.1648 | 0.0507 | 0.1383 | 0.0790 | 0.3134 |
| GU | 0.0319 | 0.0 | 0.0312 | 0.0504 | 0.0704 | 0.0648 | 0.0302 | 0.1736 | 0.0663 | 0.1117 | 0.0556 | 0.2675 |
| HI | 0.0116 | 0.0312 | 0.0007 | 0.0290 | 0.0715 | 0.0900 | 0.0190 | 0.1670 | 0.0458 | 0.1393 | 0.0705 | 0.2933 |
| MAG | 0.0414 | 0.0992 | 0.0712 | 6.3465e-06 | 0.0739 | 0.1897 | 0.0924 | 0.1710 | 0.0834 | 0.2036 | 0.1693 | 0.3491 |
| MAI | 0.0806 | 0.0875 | 0.0891 | 0.1340 | 0.0002 | 0.1394 | 0.0986 | 0.1769 | 0.0941 | 0.2168 | 0.1295 | 0.4006 |
| ML | 0.0713 | 0.0667 | 0.0773 | 0.0962 | 0.0790 | 0.0002 | 0.0695 | 0.1323 | 0.1171 | 0.0497 | 0.0403 | 0.3785 |
| MR | 0.0308 | 0.0280 | 0.0314 | 0.0503 | 0.0682 | 0.0623 | 0.0007 | 0.1625 | 0.0644 | 0.1175 | 0.0445 | 0.3423 |
| NE | 0.0949 | 0.1536 | 0.1370 | 0.1065 | 0.0955 | 0.1962 | 0.1321 | 0.0003 | 0.2130 | 0.2506 | 0.1862 | 0.3350 |
| PA | 0.0545 | 0.0935 | 0.0612 | 0.0782 | 0.0892 | 0.1573 | 0.0785 | 0.2762 | 0.0003 | 0.1716 | 0.1485 | 0.3245 |
| TA | 0.1239 | 0.1439 | 0.1384 | 0.1595 | 0.1009 | 0.0487 | 0.1204 | 0.1761 | 0.1613 | 0.0003 | 0.0972 | 0.3910 |
| TE | 0.0511 | 0.0539 | 0.0562 | 0.0785 | 0.0783 | 0.0449 | 0.0510 | 0.1513 | 0.1102 | 0.1165 | 0.0002 | 0.3401 |
| UR | 1.0 | 0.2823 | 0.5221 | 0.4771 | 0.1984 | 0.4330 | 0.4014 | 0.6438 | 0.3150 | 0.3276 | 0.5548 | 0.0001 |

**Table 15.** Cross-lingual distance between languages after applying character-level language model using perplexity-based score

| Languages | BHO | GU | HI | MAG | MAI | ML | MR | NE | PA | TA | TE | UR |
|---|---|---|---|---|---|---|---|---|---|---|---|---|
| BHO | 0.0 | 0.0381 | 0.0198 | 0.0352 | 0.0712 | 0.0860 | 0.0363 | 0.1298 | 0.0526 | 0.1311 | 0.0650 | 0.6567 |
| GU | - | 0.0 | 0.0312 | 0.0748 | 0.0789 | 0.0658 | 0.0291 | 0.1636 | 0.0799 | 0.1278 | 0.0548 | 0.2749 |
| HI | - | - | 0.0 | 0.0501 | 0.0803 | 0.0836 | 0.0252 | 0.1520 | 0.0535 | 0.1388 | 0.0634 | 0.4077 |
| MAG | - | - | - | 0.0 | 0.1040 | 0.1430 | 0.0713 | 0.1387 | 0.0808 | 0.1815 | 0.1239 | 0.4131 |
| MAI | - | - | - | - | 0.0 | 0.1092 | 0.0834 | 0.1362 | 0.0916 | 0.1589 | 0.1039 | 0.2995 |
| ML | - | - | - | - | - | 0.0 | 0.0659 | 0.1642 | 0.1372 | 0.0492 | 0.0426 | 0.4057 |
| MR | - | - | - | - | - | - | 0.0 | 0.1473 | 0.0714 | 0.1190 | 0.0478 | 0.3719 |
| NE | - | - | - | - | - | - | - | 0.0 | 0.2446 | 0.2134 | 0.1688 | 0.4894 |
| PA | - | - | - | - | - | - | - | - | 0.0 | 0.1665 | 0.1293 | 0.3198 |
| TA | - | - | - | - | - | - | - | - | - | 0.0 | 0.1068 | 0.3593 |
| TE | - | - | - | - | - | - | - | - | - | - | 0.0 | 0.4474 |
| UR | - | - | - | - | - | - | - | - | - | - | - | 0.0 |

**Table 16.** Experiments on distant language pairs.

| Model | BLEU | chrF2 | BLEU | chrF2 | BLEU | chrF2 |
|---|---|---|---|---|---|---|
| | HI → ML | | HI → TA | | HI → TE | |
| Guzmán et.al [3] | **5.12** | 30 | 7.57 | 41 | **7.19** | 26 |
| **Proposed** | 3.61 | **32** | **7.86** | **44** | 4.56 | **27** |
| | ML → HI | | TA → HI | | TE → HI | |
| Guzmán et.al [3] | 9.08 | 29 | 14.55 | 37 | 7.97 | 27 |
| **Proposed** | **9.96** | **33** | **15.43** | **40** | **9.09** | **30** |

**Table 17**. Applying on zero-shot language pairs.

| Model | HI → BHO | | BHO → HI | | HI → MAG | | MAG → HI | |
|---|---|---|---|---|---|---|---|---|
| | BLEU | chrF2 | BLEU | chrF2 | BLEU | chrF2 | BLEU | chrF2 |
| Guzmán et.al [3] | **3.34** | 14 | 4.58 | 22 | 1.67 | 13 | 4.86 | 19 |
| **Proposed** | 3.13 | **17** | **5.72** | **27** | **2.68** | **18** | **5.32** | **25** |

**Table 18**. Experiments on back-translation.

| Model | GU→HI | | | | HI→GU | | | |
|---|---|---|---|---|---|---|---|---|
| | BLEU | chrF2 | TER | WER | BLEU | chrF2 | TER | WER |
| Guzmán et.al [3] + BT(monolingual data) | 34.26 | 55 | 0.564 | 58.24 | 28.32 | 54 | 0.619 | 62.47 |
| **Proposed + BT(monolingual data)** | 35.62 | 59 | 0.554 | 57.39 | 29.29 | 58 | 0.604 | 61.73 |